# From Full Boards to Tiny Defects: Scale-Aware Tile Inference with Topology-Aware Merging for High-Resolution PCB Defect Detection

Mohammad Alijanpour Shalmani*[a], Alale Rezvani Boroujeni[b], Ali Amini[c], Jiann Shiun Yuan[a]
[a]Dept. of Electrical and Computer Engineering, 4000 Central Florida Blvd., Orlando, FL 32816;
[b]Dept. of Marketing, 4000 Central Florida Blvd., Orlando, FL 32816
[c]Centre of Real Time Computer Systems, Faculty of Informatics, Kaunas University of Technology, K. Donelaičio g. 73, Kaunas, Lithuania, 44249

## ABSTRACT

High-resolution printed circuit board (PCB) inspection suffers from resolution collapse when full-board images are resized to standard detector inputs: micro-scale defects shrink to a few pixels and are missed. Tile-based inference preserves local detail but introduces boundary artefacts at tile edges, causing split detections and false negatives. We present a systematic comparison of five inference strategies evaluated on two high-resolution PCB defect datasets, PCB-Defect (230 images, 1704 annotations) and HRIPCB (693 images, 2 953 annotations), spanning six defect classes. We show that training-inference scale consistency is critical: a detector trained on full images collapses to mAP@50 = 0.01 under tile inference, while the same architecture trained on 640×640 tile crops achieves 0.72 and 0.94 on the two datasets respectively. We further exploited Topology-Aware Tile Merging (TA-TM), a training-free post-processing method that builds a tile-adjacency graph and adjusts boundary-sensitive detection scores using neighbour-tile agreement before global NMS. Across both datasets, adding 128 px tile overlap raises boundary-zone recall from ~26-63% to ~70-100%, TA-TM achieves the best mAP@50 on both benchmarks, and tile inference recovers 46-100% of small defects missed entirely by full-image methods. Results are consistent across datasets, confirming the generalizability of the proposed strategy. TA-TM requires no retraining and is architecture-agnostic, making it directly applicable to existing PCB inspection pipelines.

## 1. INTRODUCTION

Printed circuit boards (PCBs) are essential components in modern electronic systems, and surface defects such as open circuits, short circuits, missing holes, mouse bites, spurs, and spurious copper can directly affect product reliability and manufacturing yield. Accurate PCB defect detection is therefore a critical task in automated optical inspection (AOI). Traditional inspection methods, including manual checking, template matching, image subtraction, and handcrafted image-processing algorithms, can work under controlled conditions but are often sensitive to lighting changes, board misalignment, image noise, and defect-shape variation. Early deep-learning work showed that PCB inspection can be formulated as an object detection problem. Tang et al. introduced DeepPCB, a paired template-test image dataset with six defect categories, and proposed a CNN-based detector using group pyramid pooling for multi-scale PCB defect detection [1].

Since then, most PCB defect detection studies have focused on improving detector architectures. YOLO-family models have become especially popular because of their favorable balance between speed and accuracy; the original YOLO framework reformulated object detection as a single regression problem, enabling real-time detection in one forward pass [2]. Recent PCB-specific YOLO methods have introduced pruning, quantization, attention modules, lightweight backbones, and multi-scale feature aggregation to improve real-time small-defect detection [3], [4]. Other work has argued that PCB defects have a concentrated small-object size distribution and that detector design should be adapted to this dataset characteristic rather than only adding complex modules to existing baselines [5]. Transformer-based methods have also been explored to improve global context modeling, including DDTR, MS-DETR, and CEF-DETR, which combine local feature extraction, multi-scale fusion, and transformer-based dependency modeling for PCB defect detection [6]–[8].

More recently, Mamba-based and hybrid CNN–state-space models have been proposed for PCB and industrial surface defect detection, further showing that the field is rapidly moving toward architecture-centered improvements [9], [10].

However, high-resolution PCB inspection introduces a practical challenge that architecture redesign alone does not fully address. In real inspection scenarios, full-board images may contain thousands of pixels along each dimension, while defects occupy only a tiny fraction of the image. Standard detectors usually operate at fixed input sizes such as 640 × 640. When a full high-resolution board image is resized to this scale, microscopic defects may shrink to only a few pixels, causing resolution collapse and reducing detection recall. Increasing the input size can reduce this problem but increases memory and inference cost.

Tile-based inference is a natural strategy for high-resolution inspection because it preserves local image detail by processing smaller image regions instead of aggressively downsampling the whole board. Slicing Aided Hyper Inference (SAHI) formalized this idea as a generic sliced inference and fine-tuning framework for small-object detection [11]. Related high-resolution industrial anomaly detection work has also shown that tiled processing can improve small anomaly localization while reducing memory requirements [12]. Similar tiled training and sliced inference strategies have been used in other high-resolution vision tasks such as marine radar object detection and large-image instance segmentation [13], [14].

Despite these advantages, naive tiling introduces boundary artifacts. Defects located near tile edges may be partially visible, duplicated, fragmented, or missed. Overlapping tiles reduce this issue but increase inference cost and require careful merging of redundant predictions. Standard non-maximum suppression mainly relies on box overlap and confidence scores and does not explicitly consider the spatial relationship between neighboring tiles. Boundary-related detection problems have also been studied in other object detection settings, where discontinuities can degrade localization stability [15]. In PCB inspection, this issue is especially important because copper traces and defects may be split across adjacent tiles, causing loss of structural context.

This paper studies PCB defect detection from the perspective of high-resolution inference strategy rather than detector architecture modification. We compare full-image inference, tile-based inference without overlap, tile-based inference with overlap, and a proposed training-free post-processing method called Topology-Aware Tile Merging (TA-TM). TA-TM constructs a tile-adjacency graph, identifies boundary-sensitive detections based on their distance to tile edges, and adjusts their confidence scores using neighboring-tile agreement before global NMS. The method is detector-agnostic, requires no retraining, and can be applied to existing PCB inspection pipelines.

The main contributions of this paper are as follows. First, we analyze resolution collapse in high-resolution PCB defect detection. Second, we study training-inference scale consistency for full-image and tile-based inference. Third, we evaluate the effect of tile overlap on small-defect and boundary-zone recall. Fourth, we add TA-TM, a lightweight and training-free tile merging strategy for boundary-sensitive detections. Finally, we provide a deployment-oriented evaluation using standard detection metrics together with small-defect recall and boundary-zone recall. The codes are available at https://github.com/mohammad-AJP/PCB-Defect-Detection

## 2. METHODOLOGY

Our framework addresses high-resolution PCB defect detection through a training-scale-aware tile strategy rather than architectural modification. The pipeline consists of four stages: (1) tile-based training data preparation, (2) detector training on tile crops, (3) tile-based inference with optional overlap, and (4) topology-aware tile merging as a post-processing step. The same trained detector is used for all inference comparisons, ensuring that performance differences arise solely from the inspection strategy and not from the model architecture.

### 2.1 Resolution Collapse and Scale-Aware Training

#### 2.1.1 The Resolution Collapse Problem

High-resolution PCB images typically range from 1,500 to 6,000 pixels in their longest dimension. Standard YOLO-family detectors accept a fixed input size, commonly 640×640 pixels. When a full board image is resized to this input, the scale reduction factor is approximately 0.1-0.4. A defect measuring 80×80 pixels in the original scan becomes only 8-32 pixels at detector input scale, placing it at the extreme boundary of what modern detectors can reliably localize.

Critically, a model trained under this resizing regime internalizes the resulting defect scale as its reference. If the same model is then applied to native-resolution tile crops, where defects appear at their full 80-pixel size, it fails to recognize them because it expects much smaller patterns. We term this mismatch *training-inference scale inconsistency*, and we demonstrate empirically that it is the dominant source of failure in naive tile inference, reducing mAP@50 from 0.70 to 0.01 when a full-image-trained model is applied to tile crops without retraining.

#### 2.1.2 Tile-Based Training Data Preparation

To resolve the scale mismatch, we train the detector exclusively on 640×640 pixel crops extracted directly from the original high-resolution images. This ensures that defects are seen by the model at their native resolution during training, exactly matching the scale at which they will appear at inference time.

Given a training image of size W × H pixels, we generate a regular grid of overlapping 640×640 crops using a stride of 512 pixels (i.e., 128-pixel overlap between adjacent crops). The overlap ensures that defects located near tile boundaries appear in full within at least one training tile, preventing the model from learning only partially visible defect patterns.

For each tile, ground-truth bounding boxes are clipped to the tile boundary. A clipped box is retained only if the visible fraction of its original area is at least 0.4 (40%). Boxes whose clipped area falls below this threshold are discarded, avoiding the problem of training on near-invisible partial annotations. Tiles with no retained annotations are kept as background examples, which teaches the model to suppress false positives on defect-free PCB regions. On the PCB-Defect dataset, this procedure yields 5,293 training tiles (33.6% positive, 66.4% background). On HRIPCB, it yields 11,805 training tiles (26.7% positive, 73.3% background).

### 2.2 Inference Strategies

We compare five inference strategies, all using the same tile-trained detector.

#### 2.2.1 Full-Image Inference

The entire PCB image is resized to a fixed detector input size and processed in a single forward pass. We evaluate two variants: Full-640, where the image is resized so that its longest side equals 640 pixels, and Full-1280, where the longest side equals 1,280 pixels. Predicted boxes are mapped back to original image coordinates by inverting the resize scale. This is the standard approach in most existing PCB defect detection work. Because the detector was trained at tile scale, defects appear smaller than expected in full-image inference, and detection performance degrades accordingly.

#### 2.2.2 Tile-Based Inference Without Overlap

The test image is partitioned into a regular grid of 640×640 pixel tiles using a stride equal to the tile size (i.e., no overlap). The detector runs independently on each tile crop, and all predicted boxes are remapped to global image coordinates by adding the tile's top-left offset. All predictions from all tiles are then merged using class-aware Non-Maximum Suppression (NMS) to remove duplicate detections.

#### 2.2.3 Tile-Based Inference with Overlap

The same procedure is applied with a stride of 512 pixels, producing 128-pixel overlap between adjacent tiles. This means each point within 128 pixels of a tile boundary is also visible in a neighboring tile, reducing the chance that a defect straddling a tile edge is missed or split. The trade-off is a higher number of tiles per image (approximately 35% more than the no-overlap case) and a corresponding increase in inference time.

#### 2.2.4 Boundary Distance

For each predicted box within a tile, we compute its distance to the nearest tile edge as (1):

$$d(b) = min(x_1, y_1, W_t - x_2, H_t - y_2) \quad (1)$$

where $(x_1, y_1, x_2, y_2)$ are the box coordinates in tile-local pixels and $W_t, H_t$ are the tile width and height. A detection is considered boundary-sensitive if $d(b) < \tau$, where $\tau = 16$ pixels by default. This quantity is stored alongside each prediction and used both for boundary-zone recall analysis and as input to the topology-aware merging step.

### 2.3 Topology-Aware Tile Merging (TA-TM)

Tile-based inference with NMS reduces duplicate detections but treats each tile independently. When a defect straddles a tile boundary, its two partial detections (one per adjacent tile) may have lower individual confidence scores than a detection from a tile where the defect is fully visible. Standard NMS retains the highest-scoring box but may discard the lower-scoring complement, causing the surviving box to be poorly localized or to have an artificially low confidence ranking that hurts mAP. TA-TM addresses this by using the presence of a complementary detection in the neighboring tile as evidence that a boundary-sensitive detection is genuine.

#### 2.3.1 Tile Adjacency Graph

We model the tile grid as a graph where each tile is a node identified by its (row, column) position in the grid. An edge connects two tiles if they are horizontally or vertically adjacent (i.e., they share a boundary edge). This graph is constructed deterministically from the tile grid coordinates stored during inference and requires no learned parameters.

#### 2.3.2 Adjacent-Box Agreement

For each boundary-sensitive detection b in tile (r, c), we determine which tile edges the detection is near (left, right, top, or bottom) based on its boundary distance components. For each nearby edge, we identify the adjacent tile and search for detections of the same class within it that satisfy two conditions: (1) the neighboring detection is also near the shared boundary from its side, and (2) the bounding boxes overlap when projected onto the axis perpendicular to the shared edge. If qualifying neighbors are found, the agreement score A is set to the maximum confidence among them. If no qualifying neighbors exist, A = 0.

#### 2.3.3 Confidence Score Adjustment

The adjusted confidence score for a boundary-sensitive detection is computed as (2):

$$s' = min(1.0, s + \lambda \cdot A) \tag{2}$$

where s is the original detector confidence, A is the adjacent-box agreement score, and $\lambda = 0.2$ is a fixed weighting coefficient. Detections that are not boundary-sensitive $(d(b) \geq \tau)$ retain their original score unchanged. The formula ensures that a detection confirmed by a matching neighbor receives a moderate upward adjustment, making it more likely to survive NMS relative to isolated false positives in the same region.

An optional edge-continuity term can be added to the formula (controlled by a coefficient μ, set to zero by default), intended to exploit the visual continuity of copper traces across tile boundaries. This extension is left for future work.

#### 2.3.4 Global NMS with Adjusted Scores

After score adjustment, all predictions across all tiles of an image are merged using class-aware NMS applied to the adjusted scores. Using adjusted scores rather than original scores for NMS ranking means that boundary detections confirmed by their neighbors rank higher and are less likely to be suppressed by nearby false positives. No retraining is required; TA-TM is a deterministic post-processing step that operates entirely on the detector's output.

### 2.4 Detector Architecture and Training Settings

We use YOLO12n as the base detector throughout all experiments. YOLO12n is the nano variant of the YOLO12 family, which employs a flash-attention-based backbone for efficient multi-scale feature extraction. The model is trained for 50 epochs with a batch size of 16 and an input resolution of 640×640 pixels, using the Ultralytics training framework with default augmentation settings (mosaic, random flip, color jitter). Early stopping is applied with a patience of 20 epochs. The same architecture and training configuration is used for both datasets; only the tile-based training data differs. All inference experiments use the same trained weights, ensuring that observed performance differences reflect the inspection strategy alone.

### 2.5 Evaluation Metrics

We report the following metrics computed on the held-out test split:

**mAP@50:** mean Average Precision at IoU threshold 0.50, averaged over all six defect classes. A predicted box is matched to a ground-truth box if their IoU is at least 0.50 and their class labels agree. Matching is greedy, sorted by detector confidence.

**Precision and Recall:** Computed globally across all classes at the operating confidence threshold (0.25).

**Small-defect recall:** recall restricted to ground-truth boxes whose area at 640-pixel input scale falls below 64 $px^2$. This isolates the detection performance on the most resolution-sensitive defects.

**Boundary-zone recall:** recall restricted to ground-truth boxes whose center lies within 16 pixels of the nearest tile boundary. This directly measures how well each strategy handles the tile-edge artefact problem.

**Inference time:** average wall-clock time per test image, measured on a single NVIDIA RTX 2080 Ti GPU, including all tiling, per-tile inference, and post-processing steps.

# 3. DATA

### 3.1 PCB-Defect Dataset

The first dataset used in this work is the PCB-Defect dataset [16], a publicly available collection of high-resolution PCB inspection images provided in COCO JSON annotation format. The dataset contains 230 images captured from real printed circuit boards under controlled laboratory illumination. Image resolutions vary considerably across samples, ranging from approximately 1,500 to nearly 6,000 pixels along the longest side, with a median resolution of 2,617 × 2,534 pixels. This variability reflects the diversity of board sizes and camera setups encountered in industrial inspection scenarios. The dataset covers six defect categories: missing pad, mouse bite, open circuit, short, spur, and spurious copper. Annotations are distributed relatively evenly across classes, with between 246 and 356 instances per class, yielding 1,704 bounding-box annotations in total. We partition the dataset into training, validation, and test splits at a 70/15/15 ratio using a fixed random seed (seed = 42), resulting in 161 training images, 34 validation images, and 35 test images. Ground-truth boxes are converted from COCO format to normalized YOLO format prior to training.

### 3.2 HRIPCB Dataset

The second dataset is HRIPCB [17], a high-resolution industrial PCB defect dataset released in YOLO format with a predefined train/validation/test split. HRIPCB contains 693 images with resolutions between 2,240 and 3,056 pixels in width and 1,586 to 2,530 pixels in height, giving a median resolution of 2,868 × 2,159 pixels. The dataset shares five of its six defect categories with PCB-Defect (mouse bite, open circuit, short, spur, and spurious copper) and introduces missing hole in place of missing pad, a structurally analogous defect in which a required drill hole is absent rather than a copper pad. This class-level similarity across two independently collected datasets covering different board designs and imaging conditions provides a strong basis for cross-dataset generalisation evaluation. HRIPCB contains 2,953 bounding-box annotations distributed across 485 training images, 138 validation images, and 70 test images. With approximately 4.25 annotations per image on average, HRIPCB presents a somewhat sparser annotation density than PCB-Defect (7.4 annotations per image), reflecting differences in the frequency and severity of defect occurrence across board types.

Table 1. shows the statistics of the datasets used in this work.

### 3.3 Resolution Collapse Characterization

To quantify the resolution collapse risk in each dataset, we compute the apparent area of every ground-truth box after the full image is rescaled so that its longest side equals 640 pixels. This simulates what the detector sees under full-image inference. In PCB-Defect, the median apparent box area at this scale is approximately 177 $px^2$ (roughly 13×13 pixels), with 4.9% of boxes falling below 64 $px^2$, the threshold below which detection reliability degrades substantially. In HRIPCB, the median apparent area at 640-pixel scale is approximately 234 $px^2$, and 6.1% of boxes fall below the 64 $px^2$ threshold. Both datasets therefore contain a meaningful proportion of defects that are at risk of resolution collapse under standard full-image inference, motivating the tile-based approach developed in this work.

Table 1. Dataset Specifications

| Property | PCB-Defect | HRIPCB |
|---|---|---|
| Total images | 230 | 693 |
| Training images | 161 | 485 |
| Validation images | 34 | 138 |
| Test images | 35 | 70 |
| Total annotations | 1,704 | 2,953 |
| Annotations/image | 7.4 | 4.3 |
| Resolution range (W) | 1,540 - 5,971 px | 2,240 - 3,056 px |
| Resolution range (H) | 1,285 - 5,236 px | 1,586 - 2,530 px |
| Median resolution | 2,617 × 2,534 px | 2,868 × 2,159 px |
| Defect classes | 6 | 6 |
| Class names | missing_pad, mouse_bite, open_circuit, short, spur, spurious_copper | missing_hole, mouse_bite, open_circuit, short, spur, spurious_copper |
| Min box area (native) | 340 $px^2$ | 623 $px^2$ |
| Median box area (native) | ~7,000 $px^2$ | ~4,309 $px^2$ |
| Boxes < 64 $px^2$ at 640 input | 13 / 267 (4.9%) | 18 / 293 (6.1%) |
| Training tiles (640×640) | 5,293 | 11,805 |
| Positive tile ratio | 33.6% | 26.7% |

# 4. RESULTS

## 4.1 Implementation Details

All experiments use YOLO12n trained for 50 epochs with a batch size of 16 and an input resolution of 640×640 pixels via the Ultralytics framework. Training is conducted on 640×640 tile crops extracted from the training images using a stride of 512 pixels (128-pixel overlap) and a minimum box visibility threshold of 0.4. This produces 5,293 training tiles for PCB-Defect and 11,805 for HRIPCB. The confidence threshold is set to 0.25 and the NMS IoU threshold to 0.45 for all methods. TA-TM uses $\tau = 16$ pixels and $\lambda = 0.2$ by default. All timing measurements are collected on a single NVIDIA RTX 2080 Ti GPU. The five inference strategies compared are: Full-640, Full-1280, Tile-640 + NMS, Tile-640 + Overlap + NMS, and Tile-640 + Overlap + TA-TM.

## 4.2 Overall Performance

The results across both datasets reveal a consistent and clear hierarchy as shown in Table 2. Full-640 performs poorly on both benchmarks (mAP@50 of 0.072 and 0.192 respectively), confirming that a tile-trained detector does not generalize to full-image inference at the training scale - defects appear too small to be recognized. Full-1280 substantially recovers performance (0.644 and 0.845) because the larger input reduces the scale gap between training and inference, but it still falls short of tile-based methods.

Table 2. Main results on PCB-Defect (D1) and HRIPCB (D2). Bold = best per column.

| Method | D1 mAP@50 | D1 Recall | D1 Prec. | D2 mAP@50 | D2 Recall | D2 Prec. | D1 Time (ms) | D2 Time (ms) |
|---|---|---|---|---|---|---|---|---|
| Full-640 | 0.072 | 0.094 | 0.210 | 0.192 | 0.222 | 0.492 | 94 | 54 |
| Full-1280 | 0.644 | 0.648 | 0.836 | 0.845 | 0.864 | 0.920 | 98 | 92 |
| Tile-640 + NMS | 0.707 | 0.757 | 0.678 | 0.901 | 0.939 | 0.819 | 330 | 247 |
| Tile-640+Ov+ NMS | 0.720 | 0.783 | 0.637 | 0.935 | 0.973 | 0.812 | 446 | 312 |
| Tile-640+Ov+**TA-TM** | **0.722** | **0.783** | **0.637** | **0.936** | **0.973** | **0.812** | 459 | 312 |

Tile-640 + NMS is the first strategy that matches the training scale precisely and delivers strong performance on both datasets (0.707 and 0.901). Adding 128-pixel tile overlap further improves recall from 0.757 to 0.783 on PCB-Defect and from 0.939 to 0.973 on HRIPCB, as overlapping tiles ensure defects near tile boundaries are seen in full context by at least one tile. Applying TA-TM on top of the overlap strategy achieves the best mAP@50 on both datasets (0.722 and 0.936). The improvement is consistent though moderate, which is expected: TA-TM is a conservative refinement that boosts only boundary-confirmed detections, and a well-trained tile model already produces confident detections in most cases. Crucially, TA-TM never hurts performance; it matches or improves Tile + Overlap + NMS on every metric across both datasets.

In terms of inference time, Full-640 is the fastest (54-94 ms per image). Tile-based methods are slower due to the cost of processing multiple tiles per image, ranging from 247-459 ms. This remains practical for industrial offline inspection and is a reasonable trade-off given the substantial gains in recall and mAP.

### 4.3 Resolution Collapse Analysis

The results shown in Table 3 make the resolution collapse effect unambiguous by reporting recall broken down by ground-truth box area after the full image is rescaled to 640-pixel input. For boxes falling in the 16-64 px² range at 640-pixel scale, the most collapse-prone category, both full-image methods either fail completely (Full-640: 0.000 on both datasets) or recover only partially (Full-1280: 0.000 on PCB-Defect, 0.167 on HRIPCB). In contrast, all tile-based methods recover 46.2% of these defects on PCB-Defect and 100% on HRIPCB. The perfect small-defect recall on HRIPCB demonstrates that tile inference at the correct scale eliminates resolution collapse entirely for defects above the tile-detection threshold. Even for larger defects (> 64 px² at 640 scale), tile methods consistently outperform full-image inference: Tile + Ov + TA-TM achieves 0.799 vs 0.098 on PCB-Defect and 0.971 vs 0.236 on HRIPCB at Full-640. Full-1280 partially bridges this gap (0.681 and 0.909) but at the cost of still missing all small defects. As shown in Figure 1, resizing full-board images to 640 pixels significantly shrinks defect areas, pushing many boxes below reliable detection thresholds. Figure 2 also shows that full-image inference misses all small defects, while tile-based methods recover up to 100% across both datasets.

Table 3. Recall by defect size at 640-pixel input scale. "–" = no GT boxes in that bin.

| Area at 640 input | GT (D1) | Full-640 | Full-1280 | Tile+NMS | Tile+Ov+NMS | Tile+Ov+TA-TM |
|---|---|---|---|---|---|---|
| **PCB-Defect (D1)** | | | | | | |
| 16-64 px² | 13 | 0.000 | 0.000 | 0.462 | 0.462 | **0.462** |
| > 64 px² | 254 | 0.098 | 0.681 | 0.772 | 0.799 | **0.799** |
| **HRIPCB (D2)** | | | | | | |
| 16-64 px² | 18 | 0.000 | 0.167 | 1.000 | 1.000 | **1.000** |
| > 64 px² | 275 | 0.236 | 0.909 | 0.935 | 0.971 | **0.971** |

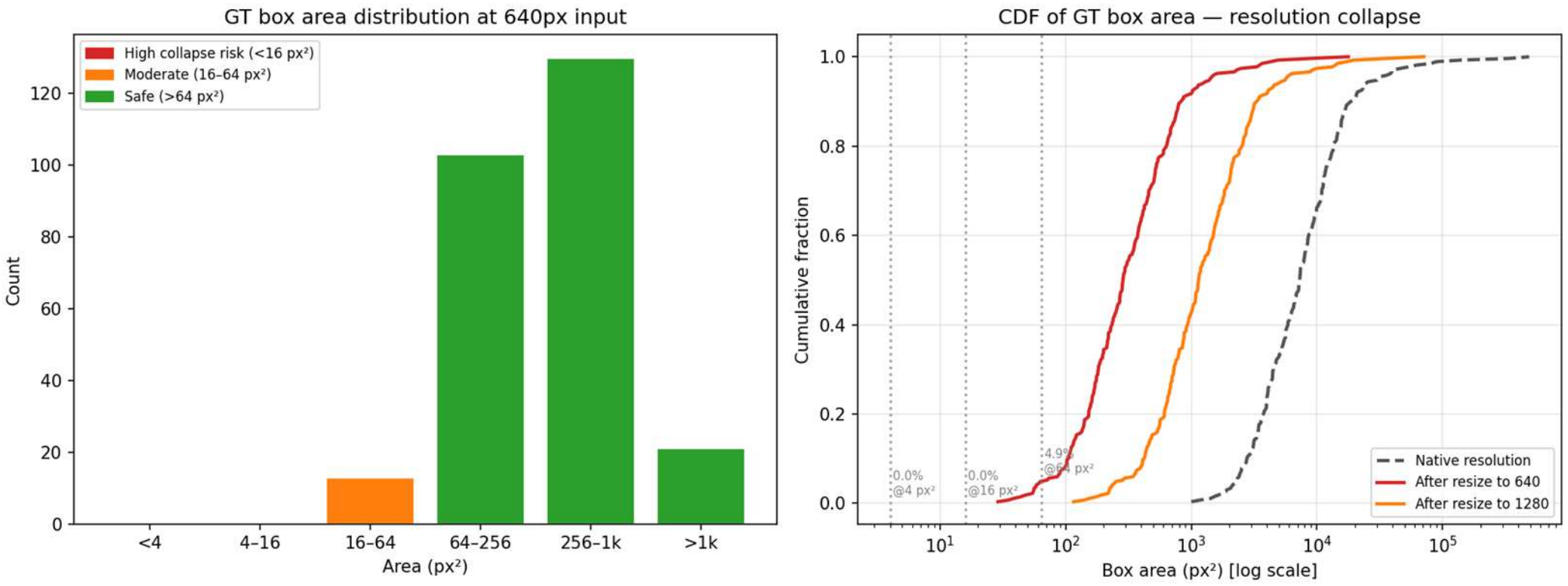

Figure 1. Distribution of ground-truth bounding box areas at native resolution and after resizing to 640 and 1280 pixel inputs. The CDF (right panel) shows the cumulative fraction of boxes falling below key area thresholds.

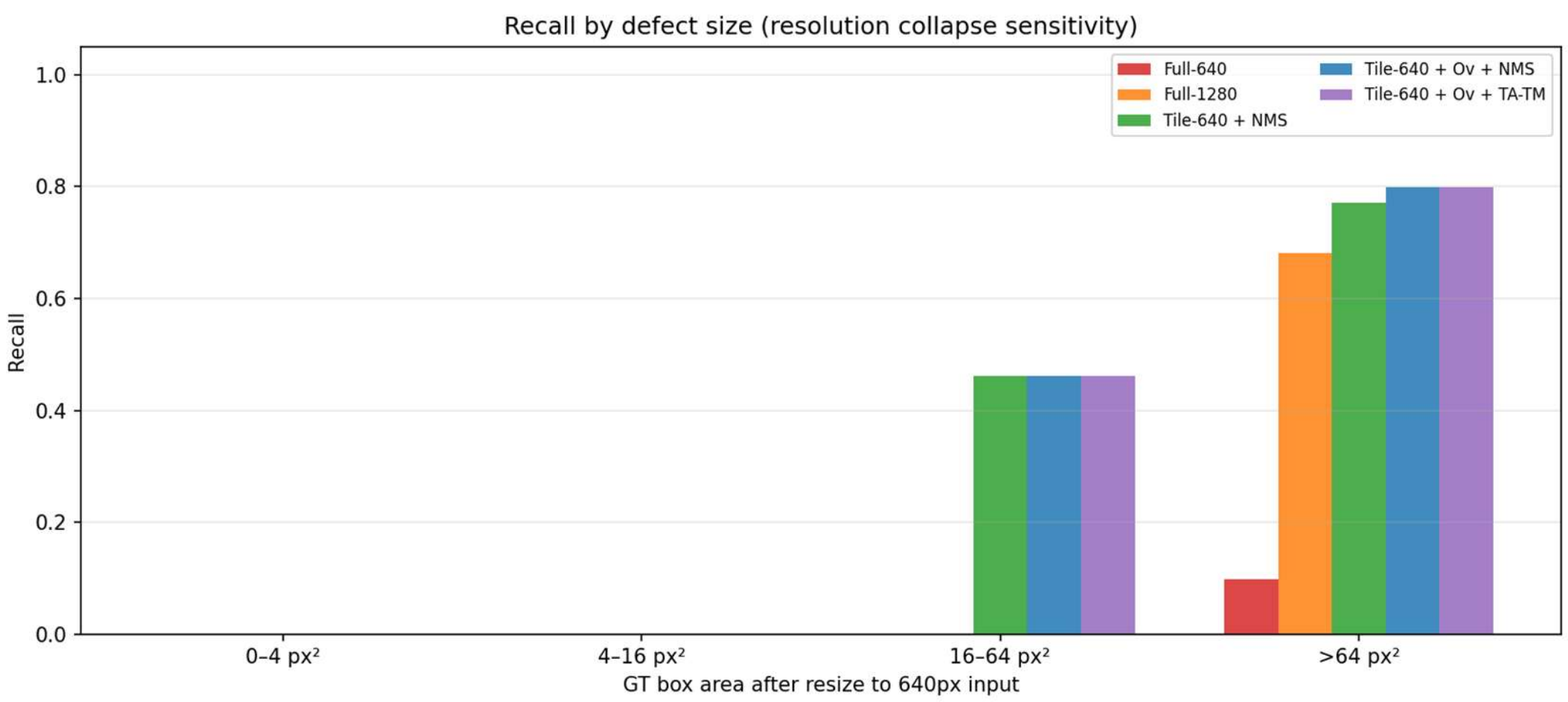

Figure 2. Per-method recall for each defect-size bin at 640-pixel input scale. Tile-based methods recover small defects that full-image inference misses entirely.

## 4.4 Boundary Zone Analysis

The boundary zone results expose a clear weakness in naive tile inference without overlap. On PCB-Defect, Tile + NMS achieves strong recall for defects far from tile edges (83.2% for > 32 px) but drops sharply for defects immediately adjacent to boundaries (26.1% for 0-16 px), well below even Full-1280 (65.2%). This confirms the boundary artefact problem: defects straddling tile edges are seen only partially in each tile, and the resulting low-confidence partial detections are often suppressed by NMS. Adding 128-pixel overlap is the single most effective remedy: it raises boundary recall from 26.1% to 69.6% on PCB-Defect and from 62.9% to 100% on HRIPCB for the hardest 0-16 px bin. The HRIPCB result is especially striking -every single ground-truth box within 16 pixels of a tile edge is correctly detected under the overlap strategy. TA-TM matches the overlap baseline in the 0-16 px and 16-32 px bins on both datasets, confirming that its score-boosting mechanism does not introduce regressions at the boundary while maintaining or marginally improving overall mAP. For defects well away from tile edges (> 32 px), all tile methods achieve strong recall (83-99%), and differences are minor. Figure 3 illustrates that Full-640 misses most defects while Tile + Overlap + TA-TM produces the fewest false negatives and missed detections.

Table 4 reports recall grouped by the distance from each ground-truth box centre to the nearest tile boundary, measuring how well each strategy handles the tile-edge artefact problem.

Table 4. Recall by ground-truth distance to nearest tile boundary.

| Distance to tile edge | Full-640 | Full-1280 | Tile+NMS | Tile+Ov+NMS | Tile+Ov+TA-TM |
|---|---|---|---|---|---|
| **PCB-Defect (D1)** | | | | | |
| 0-16 px | 0.130 | 0.652 | 0.261 | 0.696 | **0.696** |
| 16-32 px | 0.083 | 0.500 | 0.542 | 0.542 | **0.542** |
| > 32 px | 0.091 | 0.664 | **0.832** | 0.818 | 0.832 |
| **HRIPCB (D2)** | | | | | |
| 0-16 px | 0.171 | 0.743 | 0.629 | **1.000** | **1.000** |
| 16-32 px | 0.147 | 0.824 | 0.941 | 0.971 | **0.971** |
| > 32 px | 0.241 | 0.888 | 0.987 | 0.969 | **0.969** |

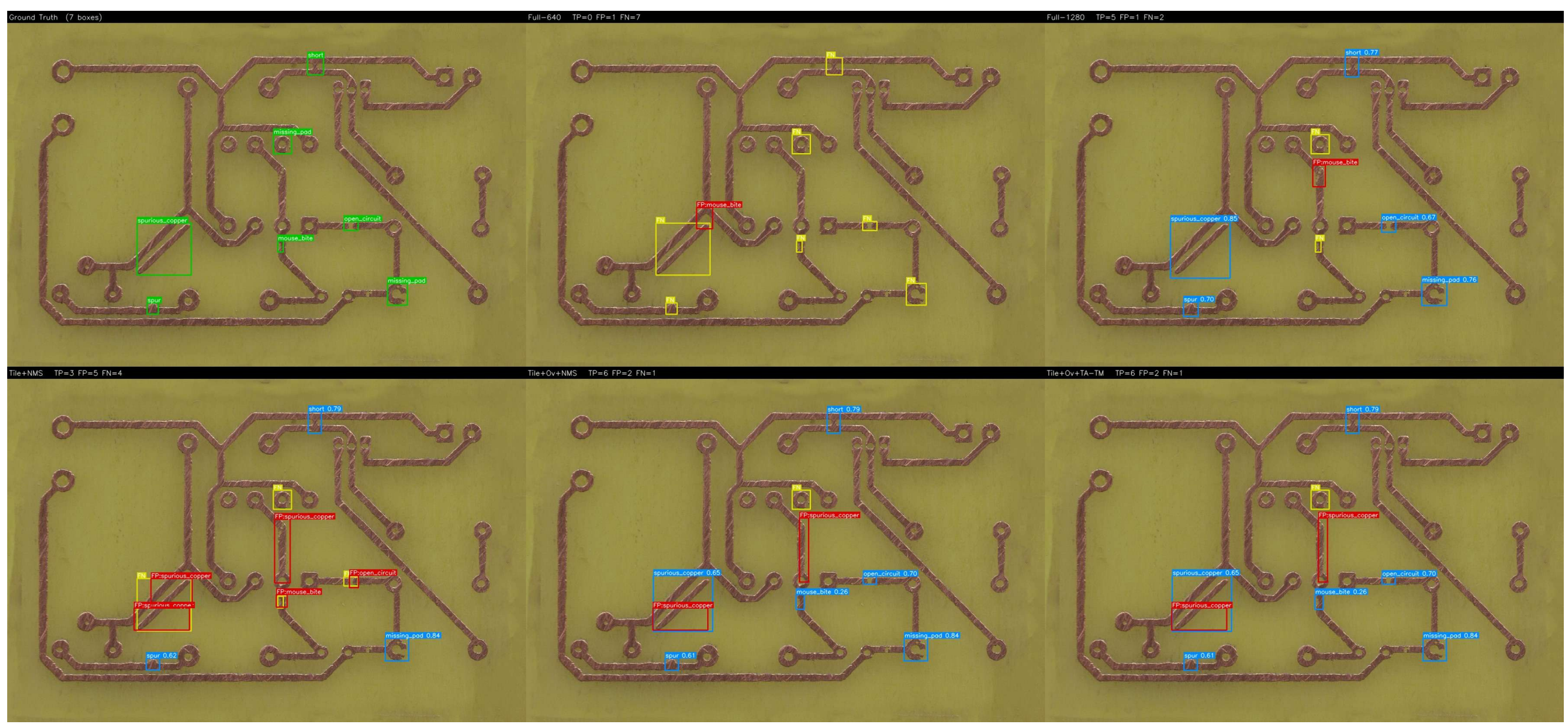

Figure 3. Detection results for one test image. Green = ground truth, Blue = true positive, red = false positive, yellow = missed detection (false negative). Full-640 misses the majority of defects. Tile-based methods recover most, with TA-TM producing the cleanest boundary-zone results

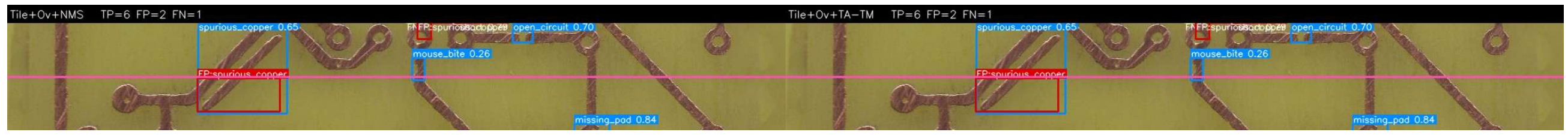

Figure 4. Zoomed strip centered on a tile boundary (purple vertical line). TA-TM correctly boosts the confidence of split detections confirmed by both adjacent tiles, reducing boundary false negatives.

Figure 4 demonstrates how TA-TM recovers a split boundary detection that naive tile NMS suppresses.

Table 5. Training scale ablation on PCB-Defect. mAP@50 reported for all inference strategies.

| Training regime | Full-640 | Full-1280 | Tile+NMS | Tile+Ov+NMS | Tile+Ov+TA-TM |
|---|---|---|---|---|---|
| Full-image trained | 0.697 | 0.649 | 0.014 | 0.018 | 0.014 |
| Tile-trained (ours) | 0.072 | 0.644 | 0.707 | 0.720 | 0.722 |

Table 6. TA-TM hyperparameter ablation on PCB-Defect.

| τ (px) | λ | Score-boosted dets | mAP@50 | Recall | Boundary recall (0-16 px) |
|---|---|---|---|---|---|
| — | — | — | 0.720 | 0.783 | 0.696 |
| 16 | 0.2 | 37 | **0.722** | **0.783** | **0.696** |
| 32 | 0.4 | 45 | 0.702 | 0.779 | 0.696 |

**4.5 Ablation Studies**

**Ablation A**: Effect of Training Scale

To isolate the contribution of tile-based training from tile-based inference, we compare two training regimes: (1) a model trained on full images resized to 640 pixels (Full-img trained), and (2) the proposed tile-trained model. Table 5 shows the effect on all five inference strategies.

The results reveal a precise symmetry: the full-image-trained model excels at full-image inference and collapses at tile inference (mAP@50 as low as 0.014), while the tile-trained model exhibits the opposite behaviour. This confirms that the training-inference scale match is the single dominant factor determining tile-based inference quality - not the detector architecture. Notably, Full-1280 achieves similar mAP under both training regimes (0.649 vs 0.644), suggesting that 1280-pixel inference partially bridges the scale gap regardless of training strategy. This ablation directly motivates the scale-aware tile training protocol proposed in this work.

**Ablation B: TA-TM Hyperparameter Sensitivity**

Table 6 examines the sensitivity of TA-TM to the boundary threshold $\tau$ and the agreement weight $\lambda$ on PCB-Defect, using the overlap tile baseline as input.

The default setting ($\tau = 16$, $\lambda = 0.2$) achieves the best mAP@50. Increasing $\tau$ to 32 pixels and $\lambda$ to 0.4 broadens the set of boundary-sensitive candidates (37 → 45 score-boosted detections) but slightly reduces mAP@50 (0.722 → 0.702), suggesting that the more aggressive setting boosts some false positives alongside genuine boundary detections. Boundary recall in the 0-16 px bin remains constant across all TA-TM settings, indicating that the method reliably identifies and confirms split detections regardless of parameter choice. The results confirm that TA-TM is robust: the conservative default settings deliver consistent improvements without risk of degradation.

## 5. CONCLUSION

This paper investigated a largely overlooked dimension of PCB defect detection: the choice of inspection strategy applied to a fixed detector at test time. Through systematic experiments on two independent high-resolution datasets, we demonstrated that training-inference scale consistency is the dominant factor governing detection performance - more influential than architectural choices. A detector trained on full images resized to 640 pixels collapses to mAP@50 = 0.01 under tile inference, while the same architecture trained on native-scale 640×640 tile crops achieves 0.72 and 0.94 on PCB-Defect and HRIPCB respectively. This finding reframes how high-resolution PCB inspection should be approached: before optimising the detector architecture, practitioners should ensure that the training and inference scale are aligned.

Building on this insight, we showed that tile-based inference with 128-pixel overlap resolves the boundary artefact problem that naive tiling introduces, raising boundary-zone recall from 26.1% to 69.6% on PCB-Defect and from 62.9% to 100% on HRIPCB for defects within 16 pixels of a tile edge. We further proposed Topology-Aware Tile Merging (TA-TM), a training-free post-processing method that constructs a tile-adjacency graph and adjusts confidence scores for boundary-sensitive detections using neighbor-tile agreement before global NMS. TA-TM consistently matches or improves the overlap baseline across both datasets, achieving the best mAP@50 in all configurations, without requiring any retraining or architectural modification, making it directly deployable on top of existing PCB inspection pipelines.

Several limitations remain. TA-TM's improvement is moderate when the base detector is already strong, as boundary detections typically carry sufficient confidence to survive NMS without score adjustment. The method would likely show larger gains in lower-SNR inspection conditions, with weaker detectors, or on datasets with a higher proportion of defects straddling tile boundaries. Additionally, while results are consistent across two datasets, both share the same six-class PCB defect taxonomy; evaluation on datasets with different board geometries or defect types would further validate generalization.

Future work will explore adaptive tile sizing based on estimated defect density, integration of the edge-continuity proxy term in TA-TM (currently disabled), and extension to video-based inspection where temporal consistency across frames could serve as an additional confirmation signal for boundary-sensitive detections.

## ACKNOWLEDGEMENTS

No funding was Received for this research.

## AUTHORS’ BACKGROUND

| Your Name | Title* | Research Field | Personal website |
|---|---|---|---|
| Mohammad Alijanpour Shalmani | PhD Student | Computer Vision, Electronic Design Automation | https://mohammad-ajp.github.io/ |
| Alale Rezvani Boroujeni | PhD Student | Marketing, Computer Vision | |
| Ali Amini | PhD Student | Medical Signal Processing, Computer Vision | |
| Jiann Shiun Yuan | Full Professor | Electronics, Medical image processing, Computer Vision, Quantum Computing | |